\documentclass[11pt]{article}
\usepackage{amssymb,amsmath,bm,graphicx}
\usepackage{mathrsfs}
\usepackage{constants}
\usepackage{enumitem}
\usepackage{textcomp}
\usepackage{url}
\usepackage{verbatim}
\usepackage{cite}
\usepackage{xcolor}
\usepackage{microtype}
\usepackage{longtable}
\usepackage{tabto}
\usepackage{amsmath}
\usepackage{float}
\usepackage{subcaption}
\usepackage{caption}
\usepackage{algpseudocode}
\usepackage{algorithm}
\usepackage{color}

\usepackage{float}

\usepackage{placeins}

\usepackage{amssymb}
\usepackage{graphicx}

\usepackage{subcaption}
\usepackage{caption}

\usepackage{csvsimple}
\usepackage[T2A]{fontenc}
\usepackage[utf8]{inputenc}
\usepackage{babel}
\usepackage{dblfloatfix}
\usepackage{amsmath}
\usepackage{lipsum}
\usepackage{amsfonts}
\usepackage{caption}

\usepackage{newtxmath,booktabs,sectsty}
\paragraphfont{\mdseries\itshape}
\usepackage{tabularx,ragged2e}
\newcolumntype{L}[1]{>{\RaggedRight\hsize=#1\hsize}X}
\newcolumntype{C}[1]{>{\Centering\hsize=#1\hsize\hspace{0pt}}X}

\date{}
\begin{document}
\newcommand{\bea}{\begin{eqnarray}}
\newcommand{\ena}{\end{eqnarray}}
\newcommand{\beas}{\begin{eqnarray*}}
\newcommand{\enas}{\end{eqnarray*}}
\newcommand{\beq}{\begin{equation}}
\newcommand{\enq}{\end{equation}}
\def\qed{\hfill \mbox{\rule{0.5em}{0.5em}}}
\newcommand{\bbox}{\hfill $\Box$}
\newcommand{\ignore}[1]{}
\newcommand{\ignorex}[1]{#1}
\newcommand{\wtilde}[1]{\widetilde{#1}}
\newcommand{\qmq}[1]{\quad\mbox{#1}\quad}
\newcommand{\qm}[1]{\quad\mbox{#1}}
\newcommand{\nn}{\nonumber}
\newcommand{\Bvert}{\left\vert\vphantom{\frac{1}{1}}\right.}
\newcommand{\To}{\rightarrow}
\newcommand{\E}{\mathbb{E}}
\newcommand{\Var}{\mathrm{Var}}
\newcommand{\Cov}{\mathrm{Cov}}
\newcommand{\Corr}{\mathrm{Corr}}
\newcommand{\dist}{\mathrm{dist}}
\newcommand{\diam}{\mathrm{diam}}
\makeatletter
\newsavebox\myboxA
\newsavebox\myboxB
\newlength\mylenA
\newcommand*\xoverline[2][0.70]{%
    \sbox{\myboxA}{$\m@th#2$}%
    \setbox\myboxB\null
    \ht\myboxB=\ht\myboxA%
    \dp\myboxB=\dp\myboxA%
    \wd\myboxB=#1\wd\myboxA
    \sbox\myboxB{$\m@th\overline{\copy\myboxB}$}
    \setlength\mylenA{\the\wd\myboxA}
    \addtolength\mylenA{-\the\wd\myboxB}%
    \ifdim\wd\myboxB<\wd\myboxA%
       \rlap{\hskip 0.5\mylenA\usebox\myboxB}{\usebox\myboxA}%
    \else
        \hskip -0.5\mylenA\rlap{\usebox\myboxA}{\hskip 0.5\mylenA\usebox\myboxB}%
    \fi}
\makeatother

\newtheorem{theorem}{Theorem}[section]
\newtheorem{corollary}[theorem]{Corollary}
\newtheorem{conjecture}[theorem]{Conjecture}
\newtheorem{proposition}[theorem]{Proposition}
\newtheorem{lemma}[theorem]{Lemma}
\newtheorem{definition}[theorem]{Definition}
\newtheorem{example}[theorem]{Example}
\newtheorem{remark}[theorem]{Remark}
\newtheorem{case}{Case}[section]
\newtheorem{condition}{Condition}[section]
\newcommand{\proof}{\noindent {\it Proof:} }

\title{{\bf\Large A correlation-based fuzzy cluster validity index with secondary options detector}}
\author{Nathakhun Wiroonsri\thanks{This author is financially supported by National Research Council of Thailand (NRCT), Grant number: N42A660991 (2023). Email: nathakhun.wir@kmutt.ac.th} and Onthada Preedasawakul \thanks{Email: o.preedasawakul@gmail.com }  \\ Mathematics and Statistics with Applications Research Group (MaSA) \\ Department of Mathematics, King Mongkut's University of Technology Thonburi}


\maketitle

\begin{abstract}

The optimal number of clusters is one of the main concerns when applying cluster analysis. Several cluster validity indexes have been introduced to address this problem. However, in some situations, there is more than one option that can be chosen as the final number of clusters. This aspect has been overlooked by most of the existing works in this area. In this study, we introduce a correlation-based fuzzy cluster validity index known as the Wiroonsri–Preedasawakul (WP) index. This index is defined based on the correlation between the actual distance between a pair of data points and the distance between adjusted centroids with respect to that pair. We evaluate and compare the performance of our index with several existing indexes, including Xie–Beni, Pakhira–Bandyopadhyay–Maulik, Tang, Wu–Li, generalized C, and Kwon2. We conduct this evaluation on four types of datasets: artificial datasets, real-world datasets, simulated datasets with second option, and image datasets, using the fuzzy c-means algorithm. Overall, the WP index outperforms most, if not all, of these indexes in terms of accurately detecting the optimal number of clusters and providing accurate secondary options. Moreover, our index remains effective even when the fuzziness parameter $m$ is set to a large value. Our R package called UniversalCVI used in this work is available at \url{https://CRAN.R-project.org/package=UniversalCVI}.

\end{abstract}




\section{Introduction} \label{sec:introduction}

 Cluster analysis is an unsupervised learning tool in machine learning that is widely used in various areas, including business, pattern recognition, data mining, medical diagnosis, and image processing, among others. It relies on the inherent properties, patterns, or similarities of objects to reveal meaningful information. The aim is to identify natural groupings within a dataset that are not initially apparent and without prior knowledge of the groups. There are several clustering algorithms, mainly categorized as centroid-based clustering (such as K-means, K-medoids, K-medians, and fuzzy c-means (FCM)), hierarchical clustering (including single linkage, complete linkage, group average agglomerative, and Ward’s criterion), density-based clustering (such as DBSCAN, DENCLUE, and OPTICS), probabilistic clustering (EM), grid-based clustering (including CLIQUE, MAFIA, ENCLUS, and OptiGrid), and spectral clustering (for more details, refer to \cite{DataClusteringBook2014} and the references therein). More recently, deep learning clustering \cite{ClusteringWithDeepLearning2018} and 3D point cloud techniques, such as PointNet, PointNet++, DGCNN, and RandLA-Net \cite{3Dpointclouds2020, 3Dpointclouds2021, 3Dpointclouds2023}, have been introduced and have garnered significant attention in this field. Recently, an alternative technique using random forest has also been introduced in \cite{rf2023}. A recent method based on both distance and density with the help of Apollonius function kernel is developed in \cite{Pou24}.

In addition to the clustering process itself, there are two associated procedures: clustering tendency assessment and cluster validation. Clustering tendency assessment is a pivotal pre-clustering task that determines the presence of clusters in a dataset and, if detected, the optimal number of clusters to be sought. On the other hand, cluster validation involves evaluating the quality and performance of a clustering algorithm on a specific dataset (see \cite{Clusteringtendency2020}). A cluster validity index (CVI) serves as a tool capable of managing both of these procedures. 

Cluster tendency is not necessary for certain algorithms like DBSCAN and OPTICS, where the number of clusters is determined automatically. However, it is essential for most clustering algorithms. In this study, our focus lies on centroid-based clustering, particularly the FCM algorithm. FCM, originally proposed by \cite{DUNN1973} and subsequently improved by \cite{FCM1984}, extends k-means by assigning a membership degree to each data point, indicating its likelihood of belonging to each group. Similar to k-means, determining the optimal number of clusters remains a prerequisite for FCM. Numerous studies have introduced and examined CVIs to ascertain the optimal number of clusters for FCM. Conventionally, the Partition Coefficient (PC) and Partition Entropy (PE), as defined in \cite{PCPE2010}, rely solely on membership degrees. Most contemporary fuzzy CVIs are grounded in concepts of intercluster and intracluster distances, compactness, and separation. These include the Xie–Beni (XB) index \cite{XB1991}, Fukuyama–Sugeno index \cite{FS1989}, Pakhira–Bandyopadhyay–Maulik (PBM) index \cite{PBM2004}, Tang index \cite{TANG2005}, Wu–Li index (WL) \cite{WL2015}, Kwon indexes (Kwon1 and Kwon2) \cite{KWON1998, KWON2021}, and Saraswat–Mittal (SMI) index \cite{SMI2021}. A unique index, the generalized C (GC) index \cite{GCI2016}, incorporates the concept of Hubert’s Gamma \cite{hubert1976} to detect the strength of the relationship between the distance of each pair of data points and its membership degree. Alongside these categories, there exist a few approaches that utilize correlation to construct CVIs. For example, Pearson correlation and Spearman’s (rho) correlation cluster validity \cite{CCV2013} compute a correlation between pairwise distances and induced partition dissimilarity. During the process of preparing this manuscript, several works have developed new indices worth mentioning. The work \cite{IMI2} introduced a new fuzzy CVI intended for imbalanced datasets.  In this study, we propose a novel correlation-based CVI as an additional option for users to choose from the extensive array of existing CVIs.

The primary objective of a CVI is unequivocally to determine the optimal number of clusters. However, certain scenarios might offer additional sub-optimal choices for selection. For example, in business, there can be multiple approaches to segment customers, or when dealing with cancer patients, multiple grouping strategies might apply. In image processing, a spectrum of optimal cluster numbers could effectively highlight the primary object within an image. This motivation drives the introduction of a novel index, termed the Wiroonsri–Preedasawakul (WP) index, designed to contribute to this field, particularly in the consistent provision of multiple optimal choices. The WP index finds its roots in the Wiroonsri index (WI), a hard CVI introduced in \cite{WIR2023}, which is exclusively compatible with k-means and hierarchical clustering. To be more precise, we propose a technique to transfer the key concept of the WI through new adjusted centroids compatible with any fuzzy clustering environment. The underlying concept is founded on a correlation between the actual distance separating a pair of data points and the distance between the adjusted centroids in relation to these points. One of the key benefits of using the original WI and the WP index is that they consistently produce multiple local optimums at different numbers of clusters. This feature allows users to choose the final number of clusters that best fits their specific applications. To the best of our knowledge, we are not aware of any works besides \cite{WIR2023} discussed this point before.

While our index varies based on the dataset and is not directly employed for cluster validation, the essential correlation component can effectively fulfill this role. Given the myriad existing CVIs, we specifically compare our index to XB, PBM, Tang, WL, GC, and Kwon2. All the CVIs except GC are selected as they are compatible with the same clustering algorithms, showcase exceptional performance, and are not tailored to specific dataset types. GC is selected since it is the most recent correlation-based fuzzy CVI we are aware of and hence it is worth to see the performance compared to ours. There are several more recent works published when we are preparing and revising our manuscript which are not in our comparison experiment. It is worth mentioning them here and we leave further analysis for future work. \cite{new2020} uses an adjustment of within-cluster distance to define a new CVI.  \cite{NZ2022} defines a new CVI using topology structure instead of centroids. \cite{WW2021} proposes a hybrid CVI by weighted other known measures. \cite{NN2022} presents a new CVI using the fuzzy set theory. \cite{MS2021, NA2022, TC2022, SN2023} introduce new CVIs intended for some specific situations such as data with missing values, and hyperellipsoid or hyperspherical data.

The rest of this work is organized as follows. Section \ref{sec:background} provides essential background information regarding FCM and existing CVIs. Our proposed index, along with its mathematical properties and complexity, are presented in Section \ref{sec:main}. Experimental results, including those from image processing, are detailed in Section \ref{sec:exp}. The concluding remarks and potential future directions are discussed in Section \ref{sec:conclusion}.

\section{Background} \label{sec:background}
In this section, we revisit the FCM algorithm and introduce six established fuzzy CVIs with which we will compare the performance of our proposed index. Let $n, c \in \mathbb{N}$, subject to the condition $c \le n$, $i \in [n]$, and $j \in [c]$, where for $k \in \mathbb{N}$, we denote $[k] = \{1, 2, \ldots, k \}$. The subsequent notations will be utilized throughout the remainder of this work.

\begin{enumerate} \label{notations}
    \item $x_i$: Data points
    \item c: Number of clusters
    \item C: Known actual number of clusters
    \item $C_j$: Set of data points in the $j^{th}$ cluster
    \item $v_j$: $j^{th}$ cluster centroid
    \item $v_0$: Centroid of the entire data
    \item $\bar{v}$: Centroid of all $v_j$
    \item $\mu = \left(\mu_{ij}\right)$: Membership degree matrix where $\mu_{ij}$ denotes the degree to which a sample point $x_i$ belongs to $C_j$.
    \item $\| x-y \|$: Euclidean distance between $x$ and $y$.
\end{enumerate}

\subsection{Fuzzy c-means}

FCM clustering is a soft centroid-based clustering technique introduced by \cite{DUNN1973}, and later refined by \cite{FCM1984}. This method involves iteratively updating $c$ centroids and the membership degree that assigns each point to each cluster, ranging from 0 to 1, until convergence is achieved or a predetermined maximum number of iterations is reached. The objective of FCM is to minimize a specific function,
\begin{equation}\label{objfunc}
    \sum_{i=1}^n\sum_{j=1}^c\mu_{ij}^m\| {x}_i-{v}_j\|^2,   
\end{equation}
where $m > 1$ denotes the fuzziness parameter. 

Initially, by randomizing centroids $v_j$, the optimization of \eqref{objfunc} is performed through the iterative update of membership degrees,
\begin{equation}\label{membership}
    \mu_{ij} = \frac{1}{\sum_{k=1}^c\left(\frac{\|{x}_i-{v}_j\|}{\|{x}_i-{v}_k\|}\right)^\frac{2}{m-1}},
\end{equation}
and the centroids,
\begin{equation}\label{centroid}
    v_{j} = \frac{\sum_{i=1}^n\mu_{ij}^m x_i}{\sum_{i=1}^n\mu_{ij}^m},
\end{equation}
for $i,j \in \{1,2,\ldots,n\}$.

\subsection{Existing fuzzy cluster validity indexes}

In this study, we assess the performance of our newly introduced index in comparison to six existing indices, which are defined as follows.

\subsubsection*{\bf Xie–Beni index \cite{XB1991}}
The XB index evaluates the compactness and separation of fuzzy c-partitions. It relies on various factors such as distances between data points and centroids, membership degree, and the minimum distance between centroids. The XB index is defined as

\begin{equation*}
  \mathrm{XB(c)} = \dfrac{\sum_{j=1}^c\sum_{i=1}^n\mu_{ij}^2\| {x}_i-{v}_j\|^2}
             {n \cdot \min_{j\neq k} \{ \| {v}_j-{v}_k\|^2 \}} .
\end{equation*}
The lowest value of XB(c) indicates a valid optimal partition.
\subsubsection*{\bf Pakhira–Bandyopadhyay–Maulik index \cite{PBM2004}}
The PBM index is defined similarly to XB, with the main difference lying in the exponent of the main term and the replacement of the minimum with the maximum. The PBM index is defined as
\begin{equation*}
    \mathrm{PBM(c)} = \left(\frac{\sum_{i=1}^n \| {x}_i-{v}_0\| \cdot \max_{j \neq k}\| {v}_j-{v}_k\|}{c\sum_{j=1}^c\sum_{i=1}^n\mu_{ij}\| {x}_i-{v}_j\|}\right)^2.
\end{equation*}
The largest value of PBM(c) indicates a valid optimal partition.
\subsubsection*{\bf Tang index \cite{TANG2005}}
The Tang index is derived from the XB index by incorporating the sum of distances between cluster centroids in the numerator. This addition addresses issues related to the monotonically decreasing trend and numerical instability. The Tang index is defined as

\text{Tang(c)}
\begin{equation*}
    \hspace{0.2cm}= \dfrac{\sum_{j=1}^c \sum_{i=1}^n\mu_{ij}^2\| {x}_i-{v}_j\|^2 + \frac{1}{c(c-1)}\sum_{j\neq k}\| {v}_j-{v}_k\|^2}{\min_{j\neq k} \{ \| {v}_j-{v}_k\|^2 \}+\frac{1}{c}}.
\end{equation*}
The smallest value of Tang(c) indicates a valid optimal partition.

\subsubsection*{\bf Wu–Li index \cite{WL2015}}
The WL index evaluates the overall compactness-separation ratio of all clusters and each individual cluster. By introducing a median factor, it addresses instability issues arising in other CVIs under the same root when two centroids are closely allocated. The Wu–Li index is defined as
\begin{equation*}
    \mathrm{WL(c)} = \dfrac{\sum_{j=1}^c\left(\dfrac{\sum_{i=1}^n\mu_{ij}^2\| {x}_i-{v}_j\|^2}{\sum_{i=1}^n\mu_{ij}}\right)}{\min_{j \neq k}\{\| {v}_j-{v}_k\|^2\} +\text{median}_{j \neq k }\{\| {v}_j-{v}_k\|^2\}}.
\end{equation*}
The smallest value of WL(c) indicates a valid optimal partition. 

\subsubsection*{\bf Generalized C index \cite{GCI2016}} The GC index is a soft version of the C-index, formulated based on relational transformations of the membership degree matrix $\mu$. It comprises four distinct variants, each with its own definition. Let
\begin{equation*}
  R_{\otimes}(\mu) = \mu \otimes\mu^T
\end{equation*}
where $\otimes$ represents one of the following matrix products: $\sim$ Sum–Product, $\sim$ Sum–Min, $\sim$ Max–Product, and $\sim$ Max–Min. Denoting $R_{\otimes}(\mu) = [r_{ij}(\mu)]$, the Generalized Hubert’s Gamma is defined as
\begin{equation} \label{gammadef}
  \Gamma_{\otimes} = \sum_{j=i+1}^n\sum_{i=1}^{n-1} r_{ij}(\mu) \cdot \| x_i-x_j \|.
\end{equation}
Moreover, let
\begin{equation*} 
  \mathrm{n_{ws}}= \left\lfloor \frac{\sum_{j=1}^c\left(\sum_{i=1}^n\mu_{ij}\right)\left(\sum_{i=1}^n\mu_{ij}-1\right)}{2} \right\rfloor.
\end{equation*}
The GC index is then defined as
\begin{equation*}
  \mathrm{GC(c)} = \frac{\Gamma_{\otimes} - min(\Gamma_{\otimes})}{max(\Gamma_{\otimes}) - min(\Gamma_{\otimes})},
\end{equation*}
where $\max(\Gamma)$ is computed similarly to \eqref{gammadef}, but the sum is taken over the first $n_{ws}$ terms after rearranging $| x_i – x_j |$ and $r_{ij}(\mu)$ in descending order, and $\min(\Gamma_{\otimes})$ is computed similarly to $\max(\Gamma_{\otimes})$, but with $r_{ij}(\mu)$ arranged in ascending order. Notably, in this study, we consider only the $(\sum\wedge)$ version, as it delivers the best results in our experiments. The smallest value of GC(c) indicates a valid optimal partition.

\subsubsection*{\bf Kwon2 index \cite{KWON2021}} 
The Kwon2 index is a generalization of the Kwon index and is designed to address three main challenges present in the original Kwon index: issues that arise when the number of clusters approaches the number of data points, numerical instability with larger fuzziness parameter ($m$), and underestimation of the true number of clusters. The Kwon2 index is defined as

\text{Kwon2(c)}

{\small
\begin{equation*}
    = \dfrac{w_1\left[w_2\sum_{j=1}^c\sum_{i=1}^n \mu_{ij}^{2^{\sqrt{\frac{m}{2}}}}  \|{x}_i-{v}_j\|^2  + \dfrac{\sum_{j=1}^c\| {v}_j-{v}_0\|^2}{\max_j \|{v}_j-{v}_0\|^2 } + w_3 \right]}{\min_{i \neq j} \| {v}_i-{v}_j\|^2 + \frac{1}{c}+\frac{1}{c^{m-1}}}
\end{equation*}
}where $w_1 = \frac{n-c+1}{n}$, $w_2 = \left(\frac{c}{c-1}\right)^{\sqrt{2}}$ and $w_3=\frac{nc}{(n-c+1)^2}$. The smallest value of Kwon2(c) indicates a valid optimal partition.

\section{Our proposed index} \label{sec:main}

In this section, we present the definition of our proposed index, establish some mathematical properties, and delve into its computational complexity.

\subsection{Definition}

Our newly introduced index draws inspiration from the recently developed Wiroonsri index \cite{WIR2023}, tailored for hard clustering methods exclusively. In the fuzzy context, however, there’s a distinction: rather than having precise knowledge of the cluster centroid occupied by each sample point, we possess only the membership degree indicating the likelihood of the sample point belonging to each cluster. This necessitates the introduction of an adjusted centroid in relation to each sample point.
Adopting the same notations as defined in Section \ref{sec:background}, let

\bea \label{newcentroid}
o_i(c,\gamma) = \frac{\sum_{j=1}^c \mu_{ij}^{\gamma} v_j}{\sum_{j=1}^c \mu_{ij}^{\gamma}}
\ena
be an adjusted centroid corresponding to the membership degree of $x_i$, and let $\gamma > 0$ signify an adapted fuzziness parameter for our index. Additionally, let
\bea \label{dmdef}
\vec{d}_v = (\| x_i-v_0 \|)_{i \in [n]}
\ena
be a vector of length $n$ encompassing the distances of all data points to the centroid of the entire dataset. Further, let
\bea \label{ddef}
\vec{d} = (\| x_i-x_j \|)_{i,j \in [n]} 
\ena
be a vector of length ${n \choose 2}$ containing distances between all pairs of data points, and
\bea \label{cdef}
\vec{\nu}(c) = (\|o_i(c,\gamma)-o_j(c,\gamma)\|)_{i,j \in [n]}
\ena 
be another vector of the same length, denoting the distances between pairs of respective adjusted centroids associated with the membership degrees of the two points. It is important to note that, throughout this work, we exclusively employ the Euclidean distance. We proceed by introducing the ensuing correlation, built upon the aforementioned adjusted centroid concept. 

\begin{definition} \label{nc}
Let $\vec{d}$ and $\vec{\nu}(c)$ be as in \ref{ddef} and \ref{cdef}, respectively. WP correlation is defined as  
\beas
\texttt{WPC}(c) = \Corr(\vec{d},\vec{\nu}(c)) 
\enas
for $c=2,3,\ldots,n$, and
\beas
\texttt{WPC}(1) = 0  \text{ \ \ or \ \ } \texttt{WPC}(1) = \frac{\texttt{SD}(\vec{d}_v)}{\max \vec{d}_v - \min \vec{d}_v}
\enas
where $\Corr(\cdot,\cdot)$ denotes a correlation coefficient. 
\end{definition}

In this work, we only consider Pearson correlation which is defined as 
\beas
\texttt{WPC}(c) = \frac{\sum_{i,j\in [n]}(d_{ij}-\bar{d})((\nu_{ij}(c)-\bar{\nu}(c)))}{\sqrt{\sum_{i,j\in [n]}(d_{ij}-\bar{d})^2}\sqrt{\sum_{i,j\in [n]}(\nu_{ij}(c)-\bar{\nu}(c))^2}}.
\enas

The user must select one of the two options for $\texttt{WPC}(1)$. If $\texttt{WPC}(1) = 0$ is chosen, then $\texttt{WPC}(2)$ often exhibits significant improvement compared to $\texttt{WPC}(1)$, frequently resulting in an optimal number of clusters at $c = 2$. On the other hand, when $\texttt{WPC}(1)$ is determined by the ratio of the standard deviation (SD) and the discrepancy between the maximum and minimum distances of each data point from the centroid of the entire dataset, it can lead to improved outcomes. The rationale is intuitive: when the SD is substantially smaller than the difference between the maximum and minimum distances, dividing the dataset into two groups can yield a more effective partition.
A substantial WPC value (close to 1) signifies a strong linear relationship between the actual distance and the distance of corresponding adjusted centroids based on the membership degrees of the two points. Subsequent subsections will reveal that WPC$(c)$ becomes significant as $c$ approaches $n$, with WPC$(n) = 1$. However, clustering $n$ observations into $n$ groups is not the purpose of performing cluster analysis. According to discussions in \cite{WIR2023}, WPI1 and WPI2 quantify the enhancements in WPC when the number of clusters is increased or decreased by one. For $c = 2, 3, \ldots, n-1$, let

{\footnotesize
\bea \label{nci1}
\texttt{WPI1}(c) &=&\frac{\texttt{WPC}(c)-\texttt{WPC}(c-1)}{1-\texttt{WPC}(c-1)}\Bigg/\frac{\max\{0,\texttt{WPC}(c+1)-\texttt{WPC}(c)\}}{1-\texttt{WPC}(c)} \nn \\ 
           &=&\frac{\left(\texttt{WPC}(c)-\texttt{WPC}(c-1)\right)\left(1-\texttt{WPC}(c)\right)}{\max\{0,\texttt{WPC}(c+1)-\texttt{WPC}(c)\}\left(1-\texttt{WPC}(c-1)\right)}
\ena
}
and 

{\footnotesize
\bea \label{nci2}
\texttt{WPI2}(c) = \frac{\texttt{WPC}(c)-\texttt{WPC}(c-1)}{1-\texttt{WPC}(c-1)} - \frac{\texttt{WPC}(c+1)-\texttt{WPC}(c)}{1-\texttt{WPC}(c)}.
\ena
}

With all the necessary components introduced, our proposed index is defined as follows.

\begin{definition} \label{nci}
Let $p \in [n-1]\backslash \{1\}$, WPC be as in Definition \ref{nc} and WPI1 and WPI2 be as in \eqref{nci1} and \eqref{nci2}, respectively. For $c=2,3,\ldots,p$, WP index is defined as

\begin{flushleft}
\textbf{Case 1}: $\max_{2\le l \le p} \texttt{WPI1}(c) < +\infty$ and $\exists l \in [p]\backslash \{1\}$ such that $|\texttt{WPI1}(l)|<\infty$.
\end{flushleft}
\beas
\texttt{WP}_p(c) =  
                 \begin{cases}
                 \scriptstyle
                     \min_{2\le l \le p} \left\{\texttt{WPI1}(l)|\texttt{WPI1}(l)> -\infty\right\}  \text{  \ \ if \ }  \texttt{WPI1}(c) = -\infty  \\
                     \texttt{WPI1}(c)   \text{ \ otherwise}.
                 \end{cases} 
\enas 

\begin{flushleft}

\textbf{Case 2}: $\max_{2\le l \le p} \texttt{WPI1}(c) = +\infty$ and $\exists l \in \{2,3,\ldots,p\}$ such that $|\texttt{WPI1}(l)|<\infty$.
\end{flushleft}

{\fontsize{9.10pt}{15pt}
\beas
\texttt{WP}_p(c) =  
                 \begin{cases}
                     \scriptstyle
                     \min_{2\le l \le p} \left\{\texttt{WPI1}(l)|\texttt{WPI1}(l)> -\infty\right\} + \texttt{WPI2}(c)  \text{  \ \ if \ }  \texttt{WPI1}(c) = -\infty \\
                     \scriptstyle
										 \max_{2\le l \le p} \left\{\texttt{WPI1}(l)|\texttt{WPI1}(l)< +\infty\right\} + \texttt{WPI2}(c)  \text{  \ \ if \ }  \texttt{WPI1}(c) = +\infty \\
										 \texttt{WPI1}(c)+ \texttt{WPI2}(c)   \text{ \ otherwise}.
                 \end{cases} 
\enas
}

\begin{flushleft}
\textbf{Case 3}:  $\forall l \in \{2,3,\ldots,p\}$, $|\texttt{WPI1}(l)|=+\infty$.
\beas
\texttt{WP}_p(c) = \texttt{WPI2}(c).
\enas
\end{flushleft}

\end{definition}

\begin{algorithm}[h!]
\caption{WP index for FCM}\label{alg:alg1}
\begin{algorithmic}
\State
\State {\textbf{Input:\hspace{0.1cm}}} $x$, $cmin$, $cmax$, $m$, $\gamma$ (default$= 7m^2/4$)
\State{\textbf{Output:\hspace{0.1cm}}} WP index where $p=cmax$ for FCM with $c$ from $cmin$ to $cmax$ 
\State \hspace{0.3cm} 1. Compute the vector of distances between all pairs of data points $\vec{d}$;
\State \hspace{0.3cm} 2. Set up the lower and upper bounds to compute WPC;
\begin{list}{}{}
    \item{\textbf{if }}{$cmin = 2$, \textbf{then} $lb = 2 $}; 
    \item{\textbf{else }}{ $lb = cmin-1$}
\end{list}
\State \hspace{0.3cm} 3. Compute WPC(c) correlation;
\begin{list}{}{}
    \item{\textbf{For }}{$c$ from $lb$ to $cmax+1$;}
    \item {\hspace{5pt} 3.1 Cluster $x$ using FCM with fuzziness $m$} 
       \item {\hspace{5pt} 3.2 Compute adjusted centroids as in \eqref{newcentroid}} 
       \item {\hspace{5pt} 3.3 Compute WPC as in Definition \ref{nc}} 
      \item{\textbf{End For }}
\end{list}
\State \hspace{0.3cm} 4. Compute $\texttt{WPI1}(c)$ and $\texttt{WPI2}(c)$ for $c$ from $cmin$ to $cmax$ 
\State \hspace{0.3cm} 5. Compute $\texttt{WP}_p(c)$ where $p=cmax$ for $c$ from $cmin$ to $cmax$
\begin{list}{}{}
    \item{\textbf{if }}{$\forall c, |\texttt{WPI1}(c)|=\infty$, \textbf{then} $\texttt{WP}_p(c)=\texttt{WPI2}(c)$}; 
    \item{\textbf{else if }}{$\forall c, \texttt{WPI1}(c)<\infty$, \\
    \textbf{then} \small $\texttt{WP}_p(c)=\max\left\{\texttt{WPI1}(c),\min_l\{\texttt{WPI1}(l)>-\infty\}\right\}$};
    \item{\textbf{else }}{\small $\texttt{WP}_p(c) = \texttt{WPI2}(c)+\min \big\{\max_l
    \{\texttt{WPI1}(l)<\infty\},$  
    $\max\left\{\texttt{WPI1}(c),\min_l
    \{\texttt{WPI1}(l)>-\infty\}\right\}\big\}$}
\end{list}
\State \textbf{Return} $\texttt{WP}_p(c)$ where $p=cmax$ for $c$ from $cmin$ to $cmax$ 
\end{algorithmic}
\label{alg:alg1}
\end{algorithm}

Clearly, the largest value of $ \texttt{WP}_p(c)$ indicates a valid optimal partition. Since our index is slightly more intricate compared to most existing ones, we provide an algorithm for computing the WP index for FCM with $c = 2, 3, \ldots, p$ in Algorithm \ref{alg:alg1}. The fuzziness parameter for our index, denoted as $\gamma > 0$, is set by default to $7m^2/4$, where $m$ is the fuzziness parameter of FCM. However, this value can be adjusted by the user. A remark below clarifies the significance of $\gamma$ and our choice of a moderate default value, which enhances the stability of our index while still maintaining a connection to the fuzziness level of the clustering method. It’s important to note that FCM can be substituted with any other clustering method that employs a membership degree matrix.

\begin{remark}
We discuss the following situations to illustrate the three cases in Definition \ref{nci} of the WP index. For simplicity, we consider $cmin$$=2$ and $cmax$$=4$.

{\bf Case 1:} Assume that $\texttt{WPC}= (0.4,0.7,0.9,0.95,0.97 )$. Then  $\texttt{WPI1}= (0.75,1.33,1.25)$. This falls into Case 1 and thus $\texttt{WP} = \texttt{WPI1}= (0.75,1.33,1.25)$

{\bf Case 2:} Assume that $\texttt{WPC}= (0.4,0.7,0.9,0.85,0.92 )$. Then  $\texttt{WPI1}= (0.75,+\infty,-1.07)$ and $\texttt{WPC2}= (-0.17,1.17,-0.97)$. This falls into Case 2 and thus $\texttt{WP} = (0.75-0.17,0.75+1.17,-1.07-0.97) = (0.58, 1.92, -2.04 )$

{\bf Case 3:} Assume that $\texttt{WPC}= (0.4,0.9,0.8,0.7,0.6 )$. Then  $\texttt{WPI1}= (+\infty,-\infty,-\infty )$ and $\texttt{WPI2}= (1.83,-0.5,-0.17)$. This falls into Case 3 and thus $\texttt{WP} = \texttt{WPI2}=(1.83,-0.5,-0.17)$ 
\end{remark}

\begin{remark}
    The range of the WP index is dataset-dependent and therefore cannot be directly employed for evaluating the quality of clustering outcomes. However, the auxiliary metric WPC, necessary for calculating the WP index, can be employed for this assessment. It’s worth noting that a WPC value approaching one signifies favorable clustering performance. 
\end{remark}

\subsection{Some properties related to the WP index}

In this subsection, we present and substantiate several properties associated with the WPC and the concept of adjusted centroids. We initiate with the following lemma, which addresses the computation of centroids and membership degrees in FCM for the specific scenario when $c = n$.

\begin{lemma} \label{cmeansconv}
For $n \in \mathbb{N}$ and $i \ne j \in [n]$, let $v_i$ and $\mu_{ij}$ be the centroid and the membership degree from the FCM algorithm with $c=n$.  Then
\begin{enumerate}
    \item $v_i = x_i$, 
    \item $\mu_{ii} = 1$ and $\mu_{ij}=0$.
\end{enumerate}
\end{lemma}
\proof

Since $c$ is set to be $n$, for any $\epsilon>0$, we can let $x_i^* = (x_{i1}+\epsilon,x_{i2},\ldots,x_{ip})$  be the initial centroids.  Then, by \eqref{membership},
\beas
\mu_{ii} = \frac{1}{\sum_{k=1}^c\left(\frac{\|{x}_i-{x}_i^*\|}{\|{x}_i-{x}_k^*\|}\right)^\frac{2}{m-1}} = \frac{1}{1+\sum_{k\ne i}\left(\frac{\epsilon}{c_{k,\epsilon}}\right)^\frac{2}{m-1}}
\enas
where $c_{k,\epsilon} = \|{x}_i-{x}_k^*\|$. As $\epsilon$ is arbitrary, by taking $\epsilon \rightarrow 0$, we have $c_{k,\epsilon} \rightarrow \|{x}_i-{x}_k\|$ for $k\ne i$, and hence $\mu_{ii} = 1$. It follows that $\mu_{ij}=0$ for $j\ne i$. Clearly, if we further update the centroids as in \eqref{centroid}, then they remain the same. 
\bbox

Next we prove the following properties of the adjusted centroids.

\begin{proposition} \label{gammathm}
Let $n,c \in \mathbb{N}$ such that $c\le n$, and $o_i(c,\gamma)$ be as in \eqref{newcentroid}. Then the followings hold.
\begin{enumerate}
    \item For any $i \in [n]$, if $\mu_{ij} \ne 0$ for all $j \in [c]$, then $\lim_{\gamma \downarrow 0} o_i(c,\gamma) = \frac{1}{c}\sum_{j=1}^c v_j = \bar{v}$.
    \item For $i\in [n]$, if $\mu_{ij} \ne \mu_{ik}$ for all $j \ne k$, then there exists $j\in [c]$ such that $\lim_{\gamma \uparrow \infty}o_i(c,\gamma) = v_j$.
    \item If $c=n$, $o_i(c,\gamma) = x_i $ for all $i \in [n]$ and $\gamma>0$.
\end{enumerate}
\end{proposition}

\proof

1) Since $\mu_{ij} \ne 0$ for all $j$,
{\small
\begin{equation*}
\begin{split}
      \lim_{\gamma \downarrow 0} o_i(c,\gamma) &= \lim_{\gamma \downarrow 0}\left(\frac{\sum_{j=1}^c\mu_{ij}^\gamma v_j}{\sum_{j=1}^c \mu_{ij}^\gamma}\right)    
    = \frac{\sum_{j=1}^c \mu_{ij}^0 v_j}{\sum_{j=1}^c\mu_{ij}^0}\\ 
    &= \frac{1}{c}\sum_{j=1}^c v_j = \bar{v}.
\end{split}
\end{equation*}
}

2) Since $\mu_{ij} \ne \mu_{ik}$ for all $j \ne k$, we can without loss of generality assume that $\max_k{\mu_{ik}} = \mu_{ij}$. Dividing the top and the bottom of $o_i(c,\gamma)$ by $\mu_{ij}^\gamma$, we have
{\small
\begin{equation*}
\begin{split}
     \lim_{\gamma \uparrow \infty}o_i(c,\gamma) &= \lim_{\gamma \uparrow \infty} \left(\frac{\sum_{k=1}^c\mu_{ik}^\gamma v_k}{\sum_{k=1}^c \mu_{ik}^\gamma}\right)   \\
     &= \lim_{\gamma \uparrow \infty}\left(\frac{ v_j +  \left(\sum_{k \ne j} \left(\frac{\mu_{ik}}{\mu_{ij}}\right)^\gamma v_k\right)}{1  + \sum_{k \ne j} \left(\frac{\mu_{ik}}{\mu_{ij}}\right)^\gamma} \right) = v_j.
\end{split}
\end{equation*}
}

3) By Lemma \ref{cmeansconv}, it is clear that
\beas
o_i(c,\gamma) = \frac{\sum_{j=1}^n\mu_{ij}^\gamma x_j}{\sum_{j=1}^n\mu_{ij}^\gamma} = x_i.
\enas
\bbox

Next we discuss some properties of the adjusted centroid in the following Remark.

\begin{remark}
\begin{enumerate}
    \item A larger $\gamma$ results in the calculation of $o_i(c,\gamma)$ with more emphasis on the largest membership degree cluster corresponding to that data point. This, in turn, leads to greater stability of the index.

    \item When $\gamma$ approaches zero, the values of $o_i(c,\gamma)$ for each data point tend to converge to the same point. Consequently, the overall stability of our index diminishes under these conditions.

    \item The special case where $\gamma = 1$ implies that $o_i(c,\gamma)$ is determined based solely on the original membership degree of the data point.

\end{enumerate}
\end{remark}

We end this subsection by stating the properties of WPC.

\begin{proposition}
For $c \in [n]$ and \texttt{WPC}$(c)$ as in Definition \ref{nc},
\begin{enumerate}
	\item $-1\le \texttt{WPC}(c) \le 1$
	\item \texttt{WPC}$(n)=1$ when applying FCM with sufficient number of iterations.
\end{enumerate}
\end{proposition}
\proof

The first item follows immediately from the fact that $\Corr$ is a correlation coefficient and that $|$WPC(1)$| \le 1$.

To prove the second item, it follows from Lemma \ref{cmeansconv} that $\vec{\nu}(n)=\vec{d}$. Therefore, WPC$(n) = \Corr(\vec{d},\vec{d})=1$. 
\bbox

\subsection{Complexity analysis}

\begin{table}[H]
\centering
\caption{CVIs complexities}
\includegraphics[width=12cm]{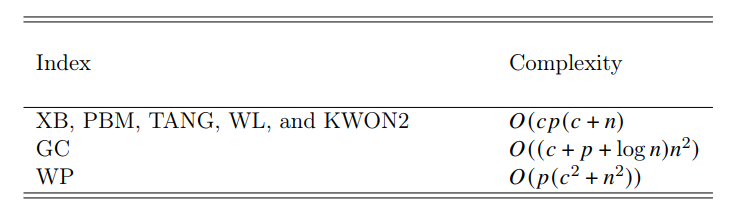}
\label{tab:complex}
\end{table}

Table \ref{tab:complex} displays the big O complexity of each index examined in this study. It is evident that our index is more complex by $O(n)$ compared to most of the existing indexes, except for GC. While WP is $O(\log n)$ faster than GC, both GC and our proposed index prove unsuitable for large-scale data. The high complexity of the WP index results from the fact that it requires computing the distances between all ${n \choose 2}$ pairs of data points. However, the subsequent section outlines how WP can be effectively applied to image processing, a field characterized by extensive data. Notably, WP not only surpasses existing indexes in accuracy but also offers certain advantages that justify its utilization.

\section{Experimental results} \label{sec:exp}
In this section, we perform two main experiments in the following two subsections. The first devotes to the sensitivity analysis of our index on its parameter $\gamma$ and the latter includes applications of our proposed index on four different types of datasets compared to others. To facilitate our experiment, we employ our dedicated R package called “UniversalCVI” (\cite{UniversalCVI}) through the RStudio environment (\cite{Rstudio}). The  “cmeans” function from the  “e1071” package (\cite{e1071}) is also employed for the computation of all the listed indexes.

\subsection{Sensitivity analysis on fuzziness parameter}

\begin{figure}[h!]
\caption{Small datasets for sensitivity analysis}
\centering\includegraphics[width=15cm]{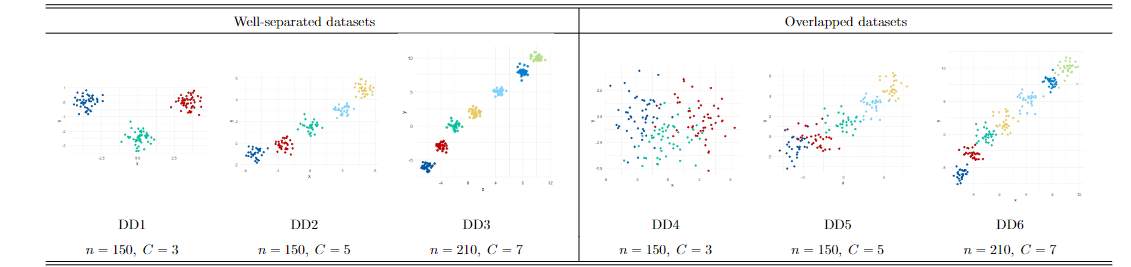}\\
\text{Note: $n$ and $C$ are the number of data points and the true number of clusters, respectively.}\\
\label{fig:gamma}
\end{figure}

In this subsection, we analyze the sensitivity of the WP index according to the fuzziness parameter $\gamma$ defined in \eqref{newcentroid} and discuss how the default parameter is set based on some additional experiments.

By Proposition \ref{gammathm}, when $\gamma$ is very small, each adjusted centroid is close to the centroid of the entire dataset. This causes all the adjusted centroids to be sensitive and close to each other which results in extremely unstable WPC and thus the WP index. On the other hand, when $\gamma$ is large, each adjusted centroid converges to one of the $c$ centroids from the FCM. This may lead to slightly high sensitivity as well in some situations, however, by this condition, it should be fitted well with well-separated datasets. Therefore, when datasets are blind, we set a moderate default parameter as $7m^2/4$ where $m$ is the fuzziness parameter of FCM. Note that we came up with this specific value by too detailed and lengthy experiment which is omitted from the paper.


\begin{table}[H]
\caption{Standard deviations of the WP index from 100 times according to $\gamma$
}
\centering\includegraphics[width=13cm]{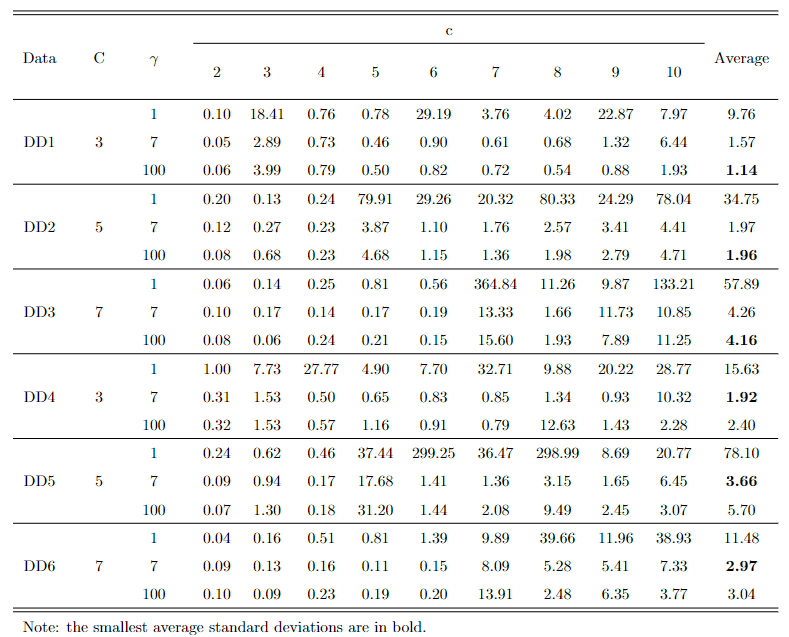}\\
\label{tab:gamma}
\end{table}

To illustrate the situation, we generate six small datasets from Gaussian distributions for 100 times each as shown in Figure \ref{fig:gamma}. The first three datasets are well-separated while the latter three are overlapped.
Then we compute the WP index for $\gamma \in \{1,7,100\}$ based on the FCM algorithm with $m=2$. Table \ref{tab:gamma} reports the standard deviations (SD) of WP$(c)$ for $c$ from 2 to 10 and their average from 100 generated datasets from the same distributions. As expected, the average SD for $\gamma = 100$ and $\gamma = 7$ are the smallest in the well-separated and the overlapped cases, respectively. Though the large $\gamma$ provides the least sensitivity in the well-separated case, the moderate one gives acceptably low sensitivity in both cases. Obviously, the small $\gamma$ provides extremely high sensitivity in all cases.

\begin{table}[H]
\caption{Sensitivity of the WP index results from 10 rounds
}
\centering\includegraphics[width=13cm]{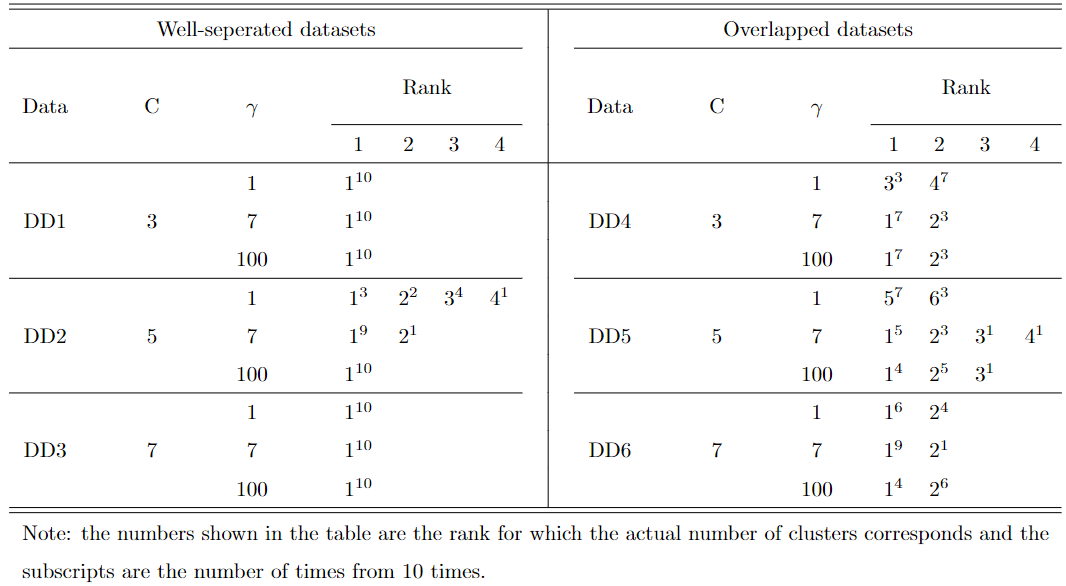}\\
\label{tab:gamma2}
\end{table}

Furthermore, we perform another experiment on the six datasets generated only once from the previous distributions. The WP index is computed for 10 times on each dataset based on the FCM algorithm with $m=2$.  In Table \ref{tab:gamma2}, we record the rank for which the actual number of clusters corresponds and the number of times it falls into that rank in the superscript. Obviously, the moderate $\gamma$ provides the best results overall and low sensitivity, though the larger $\gamma$ performs slightly better on well-separated datasets. Again, the small $\gamma$ gives the worst result.

As users do not usually know whether their datasets are well-separated or not, our default parameter is recommended and will be used in the remaining of the work.

\subsection{Applications to four types of datasets}

In this subsection, we demonstrate the efficacy of our proposed index by conducting a comprehensive experiment. Our experiment is divided into four distinct parts, each detailed in separate subsections: artificial datasets (D1–D20), real-world datasets, simulated datasets with second option (R1–R7), and image datasets (IMG1–IMG5). For our analysis, we utilize the FCM algorithm on all normalized datasets. We set the fuzziness factor to $m=1.5, 2, 4$ for the first two parts, which is well within the compatible range for most cases (as outlined in \cite{WU2012}). For the datasets with second option and image datasets, we only set $m=2$ since our main focus is to evaluate the secondary option detection performance.

Since all the datasets we consider are labeled, Table \ref{tab:acc} assesses the compatibility of the main clustering algorithm, FCM, with different values of $m$ on these datasets. Specifically, we compute the proportion of data points correctly assigned to their respective groups when setting the number of clusters, $c$, to match the actual number of groups in the dataset. We utilize our WPfuzzyCVIs package for this algorithm. It’s essential to note that if FCM demonstrates sub-optimal performance on any dataset, showcasing the efficiency of indexes based on it becomes redundant. Consequently, for a given value of $m$, we avoid considering artificial datasets with an accuracy of less than 75\%, and real-world datasets with an accuracy of less than 70\% when selecting the correct value of $c$. An overview of the results is provided in Table \ref{tab:acc}.

\begin{table}[H]
\caption{The proportion of sample points assigned into the true groups by c-means with $m=1.5, 2,4$
}
\centering\includegraphics[width=12cm]{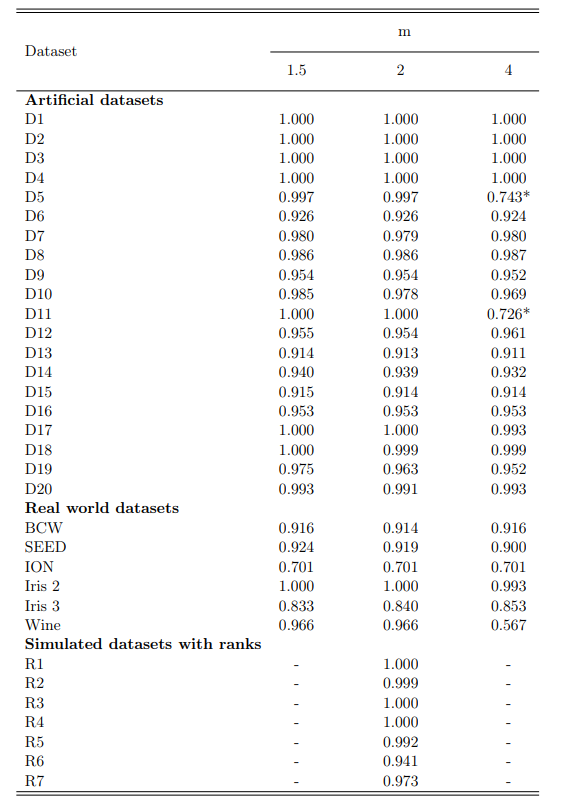}\\
\text{Note: The proportions less than $0.75$ for artificial and $0.7$ for real-world datasets}\\
\text{are marked with *}
\label{tab:acc}
\end{table}

Given FCM’s susceptibility to initial randomization, which in turn affects the sensitivity of validity indexes, we undertake a robust approach for our experiment. We execute FCM for a total of 20 rounds and select the round where the objective function in \eqref{objfunc} yields the smallest value. This approach ensures that the indexes’ performance is evaluated under conditions of minimal objective function values. Alternatively, an approach suggested in \cite{SMI2021} involves computing indexes across all rounds of FCM and selecting the most frequently occurring optimal $c$. However, we maintain our stance that if the primary clustering method yields sub-optimal results in a given round, it’s not pragmatic to compute indexes based on those results.

For comparing the efficacy of all indexes, we apply FCM to each normalized dataset, varying the number of clusters from two to $c_{\text{max}}$, where

\beas
cmax = \begin{cases}
            10  \text{  \ \ if \ } 2 \le C \le 8  \\
            15  \text{  \ \ if \ } 9 \le C \le 13 \\
            20  \text{  \ \ if \ } 14 \le C \le 18.
        \end{cases} 
\enas
Given that \cite{GCI2016} presents four alternative indexes, we specifically report on the second alternative, $C(\Sigma \wedge)$, which consistently yields the best outcomes in our experiments. Additionally, it’s important to note that the fuzziness parameter $\gamma$ for our index is set to the default value of $7m^2/4$ in all experiments conducted.

\subsubsection{Artificial datasets}

In this subsection, we focus on artificial datasets, which we classify into distinct groups based on their characteristics. These datasets include both benchmark datasets (D3, D4, D8) from \cite{Dataset2018}, as well as those we simulated ourselves (D1, D2, D9–D14, D19, D20) available in \cite{fuzzydatasets}, with additional datasets sourced from \cite{Benchmarksdata2019} (D5–D7, D15–D18). The datasets are categorized into the following groups.

\begin{list}{}{} \label{datagroup}
\item{\bf Group1: }{well-separated}{ $D1-D4$}
\item{\bf Group2: }{non-overlapped}{ $D5-D8$}
\item{\bf Group3: }{different size or density}{ $D9-D12$}
\item{\bf Group4: }{overlapped}{ $D13-D16$}
\item{\bf Group5: }{non-round shape}{ $D17-D20$.}
\end{list}

\begin{figure}[H]
\caption{Artificial datasets}
\centering\includegraphics[width=15cm]{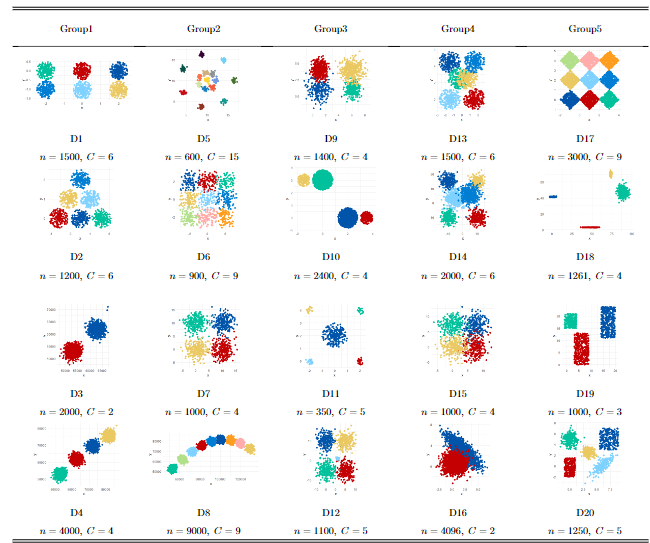}\\
\text{Note: $n$ and $C$ are the number of data points and the true number of clusters, respectively}\\
\label{fig:art_plots}
\end{figure}

The datasets’ plots and specifications are presented in Figure \ref{fig:art_plots}. Most datasets were generated from Gaussian distributions, with varying means and variances, except for D17–D20, which combine Gaussian and Uniform distributions.

To evaluate the efficiency of our proposed index in comparison to other well-known indices, we consider two criteria. Firstly, we compare the number of datasets for which each index accurately detects the known number of clusters. Secondly, we analyze the average rank to which the actual number of clusters corresponds. A well-performing index should yield an average rank close to $1$.

Table \ref{tab:art_result} illustrates the outcomes of these comparisons for various values of $m$ (1.5, 2, 4). Our proposed index demonstrates superior performance in terms of both the count of correctly detected datasets and the average rank across different $m$ values. While PBM excels for small $m$, its performance diminishes as $m$ increases. XB, Tang, and Kwon2 exhibit similar behaviors, aligning closely for $m = 1.5$ and $m = 2$, with Tang and Kwon2 slightly outperforming XB. Kwon2, as asserted in \cite{KWON2021}, emerges as the best performer among the three for $m = 4$. Kwon2 showcases compatibility with FCM for $m = 2$, yielding the second-best overall results.

\begin{table}[H]
\caption{Artificial datasets
}
\centering\includegraphics[width=15cm]{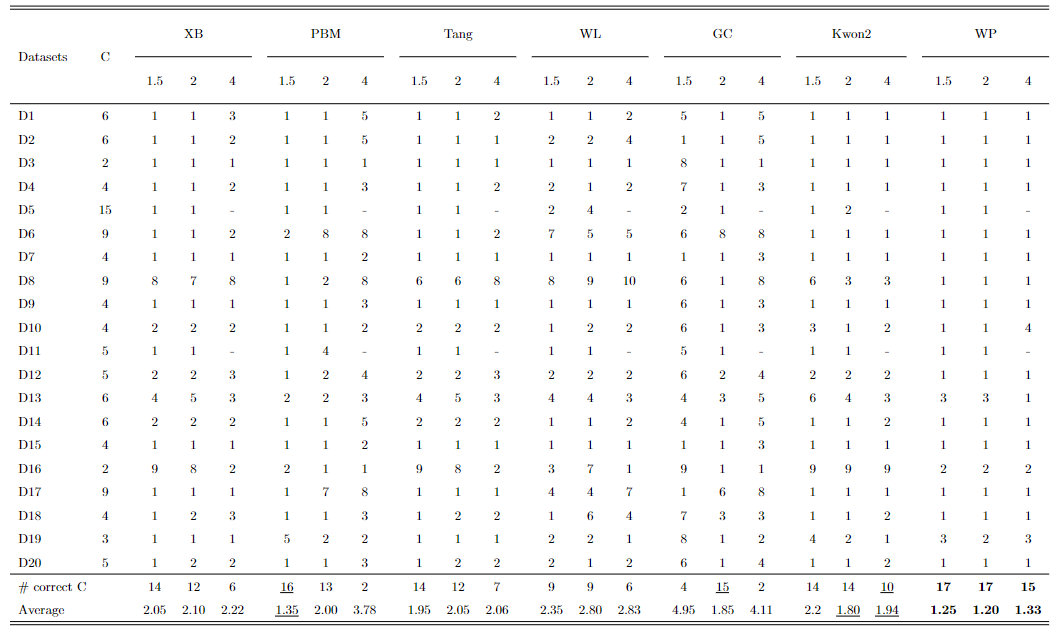}\\
\text{Note: The best and second best results are bold and underlined, respectively.}
\text{This applies to all the results tables below.}
\label{tab:art_result}
\end{table}

When assessing each index’s performance within specific dataset groups, it’s evident that existing indexes excel in well-separated datasets. On the contrary, our proposed index performs effectively across all dataset groups, with slightly lower performance in Groups 4 and 5.

\subsubsection{Real-world datasets}

\begin{table}[H]
\centering
\caption{Real world datasets speculations}
\centering\includegraphics[width=9cm]{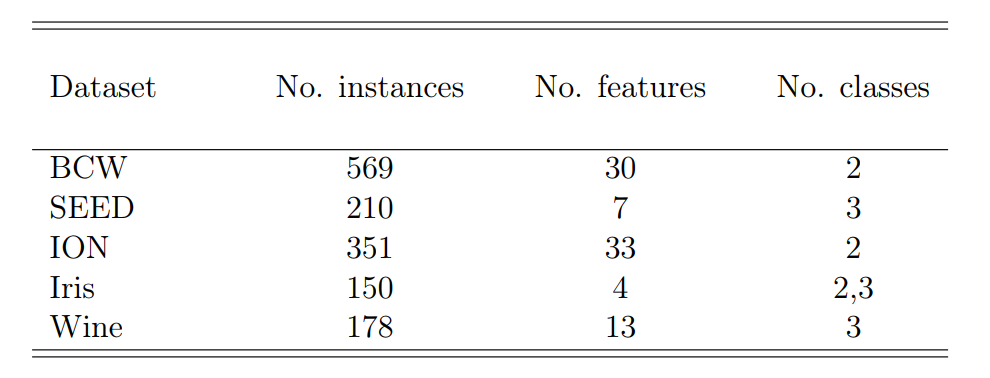}
\label{tab:UCI}
\end{table}

Moving on to real-world datasets, we analyzed five datasets from the UCI repository \cite{UCI2013}: Breast Cancer Wisconsin (BCW), Seed, Ionosphere (ION), Iris, and Wine. The specifications of these datasets are presented in Table \ref{tab:UCI}. Notably, it’s recognized that clustering the iris data can result in either 2 or 3 partitions, as indicated in \cite{KIM2004}.

\begin{table}[H]
\centering
\caption{Real-world datasets}
\centering\includegraphics[width=15cm]{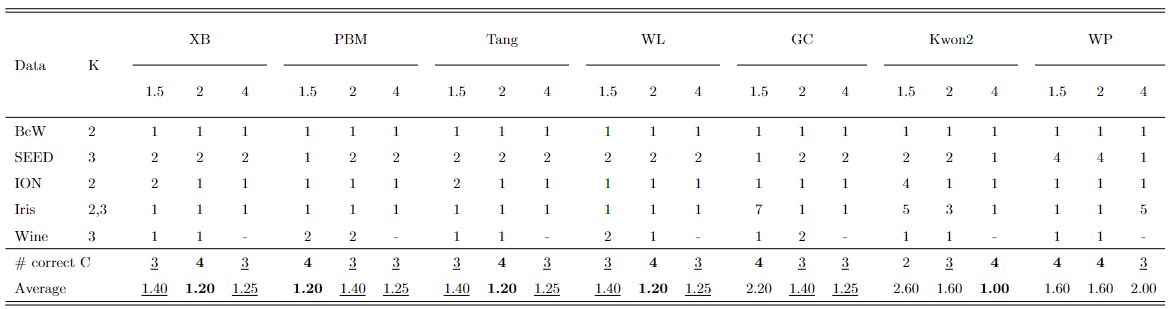}
\label{tab:real_results}
\end{table}

The experimental results are presented in Table \ref{tab:real_results}. Our proposed index demonstrates the best performance in terms of correctly detected counts for $m = 1.5$ and $m = 2$ and performs moderately well for $m = 4$. XB, PBM, Tang, WL, and GC consistently perform well for all values of $m$. Kwon2 excels particularly for $m = 4$, as claimed. It’s important to note that interpreting index performance based on real-world datasets can be challenging, as hidden subgroups might exist within labeled groups or labeled groups could be subsets of larger groups.

\begin{remark}
   We emphasize that the results of the WP index are based on the default $\gamma$ value of $7m^2/4$. For the iris data with $m = 4$, our index can correctly detect the number of clusters if we reduce $\gamma$ to 1. It’s well-known that the two groups in the iris data are overlapped, and using a larger $m$ is more reasonable. Thus, Proposition \ref{gammathm} implies that the default $\gamma$ might be too large, as it reduces the fuzziness level during index computation. However, we report the results using the default $\gamma$ as prior information about the dataset’s characteristics is often not available. If users possess prior knowledge that the data is highly overlapped or very well separated, then choosing a larger or smaller $\gamma$ accordingly can enhance the performance of our index. 
\end{remark}

\subsubsection{Simulated datasets with second option}

\begin{figure}[H]
\centering
\centering\includegraphics[width=15cm]{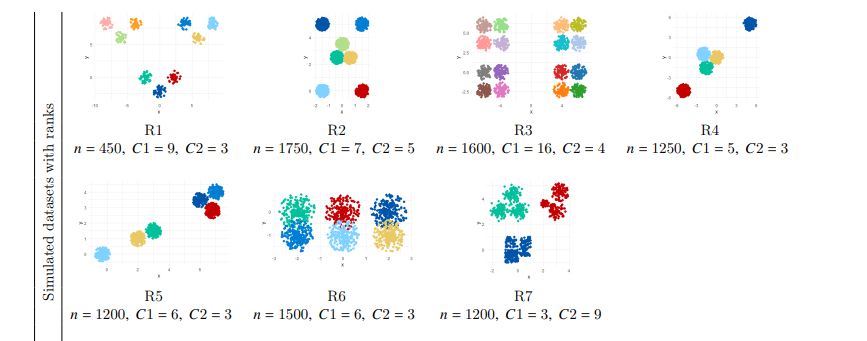}
\text{Note: $n$, $C1$, and $C2$ are the number of data points, the first and secondary options}
\text{for the number of clusters, respectively.}
\caption{Simulated datasets with second option}
\label{fig:ranked}
\end{figure}

In this subsection, we assess the performance of each index on our simulated datasets that have multiple viable options for the number of clusters. The plots of all these datasets with the first two ranks of the optimal number of clusters are displayed in Fig. \ref{fig:ranked}. We evaluate indexes based on their capability to detect both optimal and sub-optimal numbers of clusters. Table \ref{tab:ranked} shows the number of clusters detected by each index. Since we are not aware of any works considering secondary options, we establish the R-score for comparing CVI performance as the sum of sc1 and sc2 defined below, which quantifies an index’s success in detecting optimal and sub-optimal clusters. We give the first priority to the optimal number of clusters and fairly assign equal gaps between any two cases. The larger R-score indicates the better performance in detecting both options.

\beas
sc1 = \begin{cases}
            3  \text{  \ \ if the optimal number of cluster is at rank 1.}   \\
            2  \text{  \ \ if the optimal number of cluster is at rank 2.} \\
            1  \text{  \ \ if the optimal number of cluster is at rank 3.}
        \end{cases} 
\enas

\beas
sc2= \begin{cases}
            2  \text{  \ \ if the secondary number of cluster is at rank 2.}   \\
            1  \text{  \ \ if the secondary number of cluster is at rank 1 or 3.} 
        \end{cases} 
\enas

\begin{table}[H]
\centering
\caption{Simulated datasets with second option}
\centering\includegraphics[width=15cm]{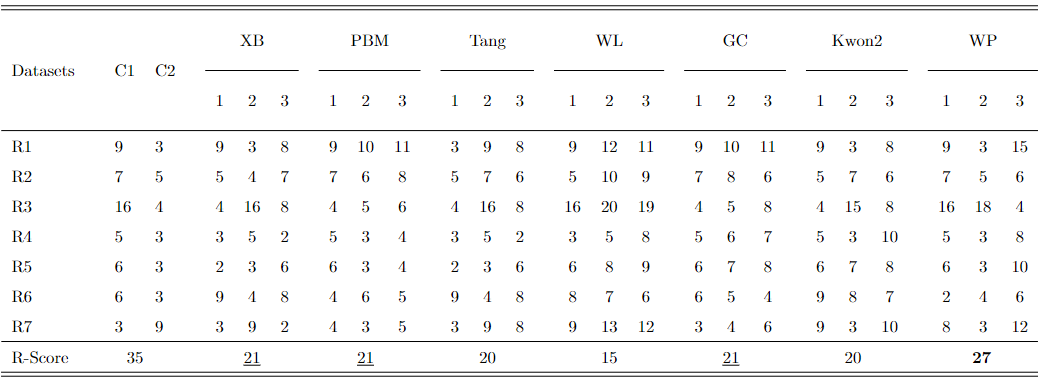}
\label{tab:ranked}
\end{table}

From Table \ref{tab:ranked}, our proposed index demonstrates strong performance in this task. It correctly detects the first and second ranks for four out of seven datasets, while for the remaining three datasets, it places the optimal number of clusters within the first three ranks.

Moreover, according to the defined score criteria, WP achieves a score of 27, outperforming XB (21), PBM (21), Tang (20), WL (15), GC (21), and Kwon2 (24).

\subsubsection{Image datasets}

\begin{figure}[H]
\centering
\caption{Image datasets}
\centering\includegraphics[width=14cm]{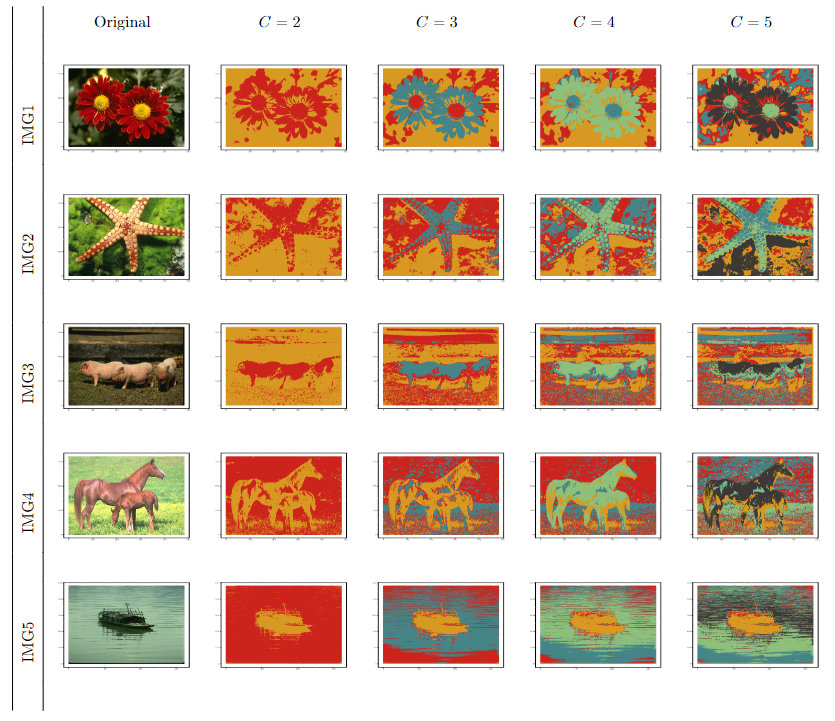}
\label{fig:image}
\end{figure}

In this subsection, we analyze five RGB-colored images from BSDS 300 \cite{BSDS300}, as shown in Fig. \ref{fig:image}. As each image is too large for GC and WP due to their complexity (see Table \ref{tab:complex}), we first reduce the image quality to 80 × 120 pixels before applying the indexes. The range of optimal numbers of clusters for each image is determined based on the clarity of the main object. For instance, in IMG1, the flower is clearly visible at $c = 2$, while $c = 3$ and $c = 4$ provide more detail in the pollen area. Similarly, $c\in [2,4]$ is acceptable for IMG2 to capture the starfish, while higher values introduce noise. For the last three images, $c > 2$ increases unnecessary noise around the main objects, making $c = 2$ the most reasonable choice. As image analysis is inherently subjective, we state only ranges without explicit rankings. Again as we are not aware of any works considering secondary options in the context of image clustering, we self-establish the I-Score criterion as

\beas
ISC = \begin{cases}
            3  &\text{  \ \ if $r_i \in A$ for all $i$}   \\
            2.5  &\text{  \ \ if $r_i \in A$ for only $i=1,2$} \\
            2  &\text{  \ \ if $r_i \in A$ for only $i=1,3$} \\
            1.5  &\text{  \ \ if $r_i \in A$ for only $i=1$ or $i=2,3$} \\
            1  &\text{  \ \ if $r_i \in A$ for only $i=2$} \\
            0.5  &\text{  \ \ if $r_i \in A$ for only $i=3$}
        \end{cases} 
\enas
where $A$ represents the set of acceptable numbers of clusters, and $r_i$, $i=1,2,3$, represent the first three ranks from an index, respectively.
To be more precise, we rank the situations from best to worst in detecting hidden number of clusters within an image dataset. Then we fairly assign equal gaps between any two cases. Clearly, the larger I-score indicates the better performance on an image dataset.

\begin{table}[H]
\centering\includegraphics[width=15cm]{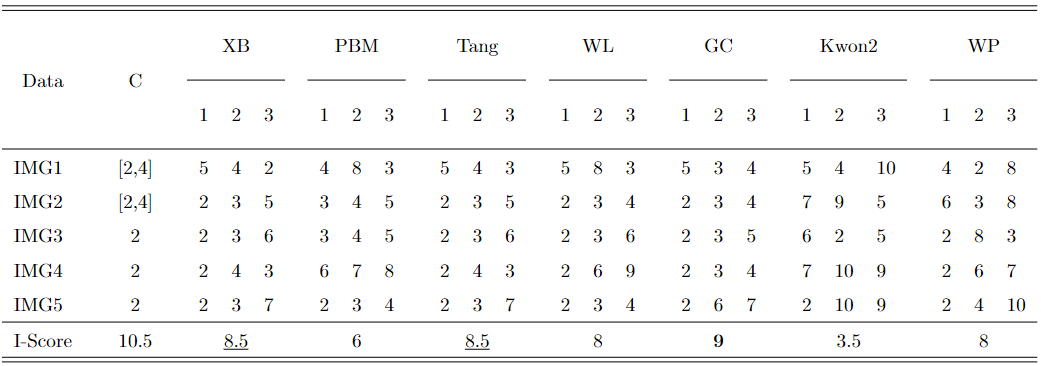}
\caption{Image datasets}
\label{tab:image}
\end{table}

The results are shown in Table \ref{tab:image}. Among the indexes, GC scores the highest with 9 out of 10.5 points. XB, Tang, WP, and WL also perform well, achieving scores of 8.5, 8.5, 8, and 8, respectively.

\subsubsection{Analysis of the WP index}

Based on our experiments on the four types of datasets, the WP index has the main advantages that its performance remains stable for more complicated datasets as in Table \ref{tab:art_result} and it provides the best result in detecting secondary options when the two options are not consecutive as shown in Table \ref{tab:ranked}. However, as in Table \ref{tab:image}, it detects mostly the first option when all the options are successive as in the image data case. The main disadvantage of our index is the computational time as discussed earlier and it requires some undersampling when handling large datasets including image datasets. This may result in a slightly worse performance.

\section{Conclusion} \label{sec:conclusion}
In this study, we have introduced the WP index, inspired by the Wiroonsri index. Unlike the original index, our proposed WP index is potentially applicable to any soft clustering method that provides membership degrees. However, the performance has been tested only on the FCM algorithm. It is defined based on the correlation between the actual distance between a pair of data points and the distance between adjusted centroids considering the membership degrees. 
The primary motivations behind developing this new index and introducing it as an option among the existing fuzzy cluster validity indices are as follows:
\begin{enumerate}
  \item \textbf{Unique concept and precision:} WP index operates under a distinctive concept compared to other existing indices, yet it exhibits a high degree of precision in detecting the optimal number of clusters.
  \item \textbf{Sub-optimal detection:} WP index offers the advantage of identifying sub-optimal numbers of clusters, allowing users to make personalized selections. 
  \item \textbf{Flexible parameter:} WP index has a fuzziness parameter $\gamma$ where users can select, though we recommend to use our default value.
\end{enumerate}
We compare our proposed index with XB, PBM, Tang, WL, Generalized C, and Kwon2 indexes across four types of datasets: artificial datasets, real-world datasets, simulated datasets with second option, and image datasets. We use the FCM algorithm with fuzziness parameter values ranging from 1.5 to 4. For the first two types of datasets, we assess performance based on the number of correctly identified optimal clusters and the average rank across all datasets where the optimal cluster numbers are applicable. For the last two dataset types, we establish new scoring criteria based on the first three options provided by each index. Our evaluation approach and results are summarized as follows.
\begin{enumerate}
  \item Among the 20 artificial datasets, the WP index clearly demonstrates superior performance in all aspects.
  \item Across the five real-world datasets, the WP index stands out as one of the top performers when the fuzziness parameter $m$ is set to 1.5 and 2, and it shows moderate performance for $m = 4$.
  \item Among the seven simulated datasets with second option, the WP index outperforms the other indices based on our scoring criterion.
  \item Across the five image datasets, the WP index achieves moderate performance according to the defined criterion.
\end{enumerate}

We also conduct a sensitivity analysis of the WP index based on the fuzziness parameter $\gamma$. Our default parameter of $7m^2/4$ provides low sensitivity in most cases and has the best performance overall. Though a larger $\gamma$ is slightly more suitable in the well-separated case, it maybe hard to know whether datasets are well-separated or not in reality. There maybe some other values of $\gamma$ which are more appropriate in different and more complicated contexts, nevertheless, using our default value is recommended as datasets are usually blind especially in higher dimensional. Analyzing this parameter under different and more complicated contexts would be an interesting future research which will benefit users with some insights on their datasets.

The primary concern with the WP index is its reliance on computing distances between all pairs of data points, rendering it unsuitable for direct application to large datasets. It is important to note that due to this limitation, it is not compatible with big data scenarios. However, we put forward a potential solution for handling big data in the future: by initially undersampling a large dataset and then applying the WP index. While this approach has not been thoroughly tested yet, preliminary results suggest its effectiveness, particularly evident when we deliberately lower the image quality in our image data experiment. Enhancing the index’s performance within this context and comparing it with other recent CVIs mentioned in the introduction present an unexplored avenue for future research.

\end{document}